\pdfoutput=1
\documentclass[11pt]{article}
\usepackage{lipsum}

\usepackage{graphicx}
\usepackage{caption}
\usepackage{subcaption}
\usepackage{enumitem}
\usepackage{natbib}
\usepackage{caption}
\usepackage{tikz}
\usepackage{amssymb}
\usepackage{xcolor}         
\usepackage{colortbl}         
\usepackage{amssymb}

\usepackage{emnlp}
\usepackage{multirow}

\usepackage{times}
\usepackage{latexsym}
\usepackage{float}
\usepackage[T1]{fontenc}

\usepackage[utf8]{inputenc}

\usepackage{microtype}

\usepackage{inconsolata}

\title{Syntax-Guided Transformers: Elevating Compositional Generalization and Grounding in Multimodal Environments}

\author{Danial Kamali \\
  Michigan State University\\
  \texttt{kamalida@msu.edu} \\\And
  Parisa Kordjamshidi \\
  Michigan State University\\
  \texttt{kordjams@msu.edu} \\}

\begin{document}
\maketitle
\begin{abstract}
Compositional generalization, the ability of intelligent models to extrapolate understanding of components to novel compositions, is a fundamental yet challenging facet in AI research, especially within multimodal environments. In this work, we address this challenge by exploiting the syntactic structure of language to boost compositional generalization. This paper elevates the importance of syntactic grounding, particularly through attention masking techniques derived from text input parsing. We introduce and evaluate the merits of using syntactic information in the multimodal grounding problem. Our results on grounded compositional generalization underscore the positive impact of dependency parsing across diverse tasks when utilized with Weight Sharing across the Transformer encoder. The results push the state-of-the-art in multimodal grounding and parameter-efficient modeling and provide insights for future research.

\end{abstract}

\section{Introduction}
Compositional Generalization refers to the ability of an intelligent agent to generalize its understanding of the underlying structure of a problem, especially when it is faced with novel compositions of the previously seen building blocks or components \cite{Chomsky57a, MONTAGUE}. It is fundamental for models to be able to extrapolate from their training environment to novel situations, a common occurrence in real-world applications. \citet{Hupkes2019} categorizes compositional generalization capabilities into five categories, \textit{systematicity},\textit{ substitutivity}, \textit{localism \& globalism}, and \textit{overgeneralization}. These abilities are crucial for models to achieve strong performance on tasks that require reasoning and understanding of hierarchical structures, such as natural language understanding, object classification, and robotics. 


Humans understand new compositions of previously observed
concepts and simpler constructs. On the other hand, despite remarkable progress in the field of Artificial Intelligence, even state-of-the-art language models demonstrate limitations in this aspect \cite{lake2018generalization,85d395469caa4cc383259d3012f528bc,shaw-etal-2021-compositional}. Especially, they often fail to effectively generalize in the reasoning depth, which involves handling multi-turn reasoning about entities and their properties in the world or even the co-occurrence of unseen spatial relations \cite{Wu-ReaSCAN}. These limitations indicate a crucial need for innovative approaches to address these issues.


\begin{figure}[t]
    \centering
         \includegraphics[width=0.4\textwidth]{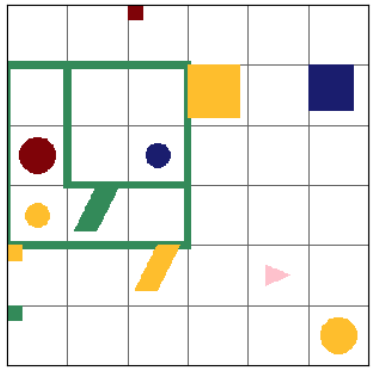}\\
         \raggedright
         \textbf{Input Command:} pull the small blue object that is inside of the small green box and in the same row as the red circle while zigzagging.\\ 
         \textbf{Action sequence:} \textit{turn left, turn left, walk, turn right, walk, turn right, walk, pull}
         \caption{This example is taken from the ReaSCAN dataset. Here, an agent is provided with a command. Its objective is to generate/execute a series of predefined actions to fulfill the task within the given environment.}
         \label{fig:reascan}
\end{figure}




In this research, our objective is to exploit the syntactic structure of language to enhance compositional generalization. Our focus is mainly on the multimodal problem setting that entangles vision and language. In this unique setting, compositional linguistic descriptions must be accurately grounded in the environment to devise coherent action plans or achieve specific goals. An illustrative example of this scenario is shown in Figure \ref{fig:reascan}.

The motivation behind leveraging syntax in our approach stems from the inherent structure and compositionality of natural language. Syntactic parsing provides crucial structural information about how words in a sentence relate to each other. We hypothesize that syntactic structure can improve intelligent agents' ability to discern the applicable attributes and descriptions for each object in its environment and better apprehend deeper levels of reasoning.

By imposing an understanding of language structure through syntactic parsing, we aim to extend the capabilities of current multimodal language models. This could potentially pave the way for more sophisticated models capable of robustly interacting with dynamic and complex vision and language environments. Apart from using structure, we equipped our end-to-end model with weight sharing that has demonstrated improving the generalization capabilities in single-modality tasks. 

 As a result, we reach state-of-the-art performance on the ReaSCAN compositional generalization benchmark, showing improvement across all test splits, especially ones requiring sentence structure comprehension. In summary, our contributions include:  



\begin{itemize}
\item  Enhancing grounded compositional generalization by integrating syntactic parsing into our model. 

\item Using syntax-guided attention masking along with weight sharing, we build a highly parameter-efficient model compared to baselines. 

\item Our model has shown marked improvement in performance across a variety of tasks that are designed for compositional generalization evaluation while enhancing computational efficacy. 

\end{itemize}

\section{Related Work}

The machine learning research community primarily focused on understanding the error bounds and the bias-variance trade-off~\cite{Hastie2009} to understand and improve the models' generalizability. Later, techniques like dropout~\cite{dropout} were introduced to improve neural models' generalization. Recently, studies have examined the generalizability of various neural network architectures using specialized generalization evaluation tasks~\cite{Hupkes2019,ontanon-etal-2022-making,csordas-etal-2021-devil}. Additionally, numerous datasets such as SCAN \cite{lake2018generalization}, CFQ \cite{Keysers2019}, and COGS \cite{cogs} have been developed to assess compositional generalization capabilities. Diverse strategies such as data augmentation \cite{andreas-2020-good, shaw-etal-2021-compositional}, innovative architectural designs \cite{korrel-etal-2019-transcoding, gao-etal-2020-systematic}, and neuro-symbolic methods \cite{Mao2019NeuroSymbolic}, have been proposed to enhance these capabilities. Consequently, these advances in text-based generalization have inspired research in multimodal compositional generalization, with developments including complex benchmarks like gSCAN \cite{ruis2020benchmark} and ReaSCAN \cite{Wu-ReaSCAN}, and advanced architectures applied to multimodal grounding \cite{kuo-etal-2021-compositional-networks,jiang-bansal-2021-inducing,qiu-etal-2021-systematic, Sikarwar2022,shaw-etal-2021-compositional}.

Furthermore, recent research highlights the significant role of syntactic information in enhancing neural models' compositional generalization capability. \citet{kuo-etal-2021-compositional-networks} suggested aligning the compositional structure of networks with the problem domain, resulting in a dynamic compositional neural network. Moreover, \citet{shaw-etal-2021-compositional} and \citet{qiu-etal-2022-improving} recommended grammar induction-based data augmentation techniques to improve compositional generalization. Unlike our work that focuses on input command structure,~\citet{9340248} introduced the concept of using parse tree node annotations in the target sequence of sequence-to-sequence tasks for enhancing compositional generalization. Meanwhile, \citet{kim2021improving} incorporated parse tree nodes into the ETC~\cite{ainslie-etal-2020-etc} model. They employed attention masking specific for ETC to symbolize the relations of tokens and aid this model in a simplified classification task based on the CFQ dataset.

We are inspired by previous research ~\cite{kim2021improving} that employs a similar technique with manually extracted parses for compositional generalization on the single text modality. However, our model utilizes off-the-shelf parsers instead of accurate manually generated parse trees, and it is generally applicable independent of the underlying models.

\section{Problem Setting}
Various studies on compositional generalization have presented a range of tasks and problem settings \cite{lake2018generalization, Keysers2019, cogs, Wu-ReaSCAN, ruis2020benchmark}.
These datasets are comprised of a training set and several test sets. To ensure rigorous evaluation, the test sets have been deliberately structured to differ from the training set in a way that requires the compositional capability to succeed.
Our paper focuses on grounding natural language instructions in the visual modality, where we map words to specific objects or actions in a multimodal environment that provides a framework to evaluate an intelligent agent's compositional structures and spatial reasoning capabilities.

We use the most recent multimodal compositional generalization benchmarks to assess our models comprehensively. In these benchmarks, an agent receives natural language instruction to carry out an action or navigate specific environments. These datasets are inherently synthetic, and they have been carefully crafted to guarantee that the test sets are systematically different from the training sets. By placing commands within a spatial context, these benchmarks bridge the gap between abstract cognitive understanding and practical action execution. Consequently, they stand as both a scholarly tool for studying compositional generalization and a valuable resource for fields like robotics that require comprehension of spatially anchored commands.

Among these benchmarks, our primary focus is ReaSCAN, owing to its heightened complexity and recent introduction to the academic community. An example of this dataset, depicted in Figure~\ref{fig:reascan}, consists of three main components: The initial state of the world, the provided input command, and the corresponding target command. Tasked with this information, the agent aims to infer the target command by leveraging both the information from the input command and the initial state.
Structurally, the world's representation in ReaSCAN is formulated as a  6×6×17 matrix. Each matrix cell comprises a 17-dimensional vector encapsulating information pertaining to an object's attributes—namely, color, shape, and size—along with indicators of the agent's positioning and orientation. The evaluation metric for this dataset is the percentage of exact matches of the predicted action sequence.
The ReaSCAN dataset includes one random test split that mirrors the training's component and compound distribution, in addition to seven compositional generalization test splits. Each of these splits is designed to probe a specific facet of a model's grounding generalization capability, as detailed in Table \ref{table:reascan}. Category A test splits delve into novel attribute compositions at both the command and object levels, drawing inspiration from gSCAN. Category B test splits assess the model's ability to generalize to unprecedented co-occurrences of concepts and spatial relations. Meanwhile, Category C probes the model's capacity to extrapolate from simple command structures to more intricate structures with higher levels of reasoning~\cite{Wu-ReaSCAN}. To illustrate the A1 split, all examples with commands containing variations of "yellow square" (such as "small yellow square" or "big yellow square") are excluded from the training data. This prevents models from associating targets with that phrase. However, the training set does include examples like "yellow cylinder" and "blue square." As a result, during testing, models are expected to accurately interpret the "yellow square" even without prior exposure to the actual composition.

\begin{table}[t]
\centering
\begin{tabular}{ll}
\hline
Split & Held-out Examples \\
\hline
Random&\footnotesize  Random.\\
A1&\footnotesize  \textit{yellow square} referred with color \& shape.\\
A2&\footnotesize  \textit{red square} referred in the command.\\
A3&\footnotesize  \textit{small cylinder} referred with size and shape\\
\multirow{2}{*}{B1}&\footnotesize co-occur of \textit{small red circle} and \textit{big blue} \\
& \footnotesize \textit{square}.\\
 
\multirow{2}{*}{B2}&\footnotesize  co-occur of \textit{same size as} and \textit{inside of} \\ & \footnotesize relations.\\
\multirow{2}{*}{C1}&\footnotesize Additional conjunction clause depth added \\
    &\footnotesize to \textit{2-relative-clause commands}.\\

\multirow{2}{*}{C2}&\footnotesize \textit{2-relative-clause} command with \textit{that is}\\
&\footnotesize instead of \textit{and}.\\

\hline
\end{tabular}
\caption{ReaSCAN dataset test splits.}
\label{table:reascan}
\end{table}

\section{Proposed Method}
\begin{figure*}[t!]
    \centering
    \includegraphics[width=\textwidth]{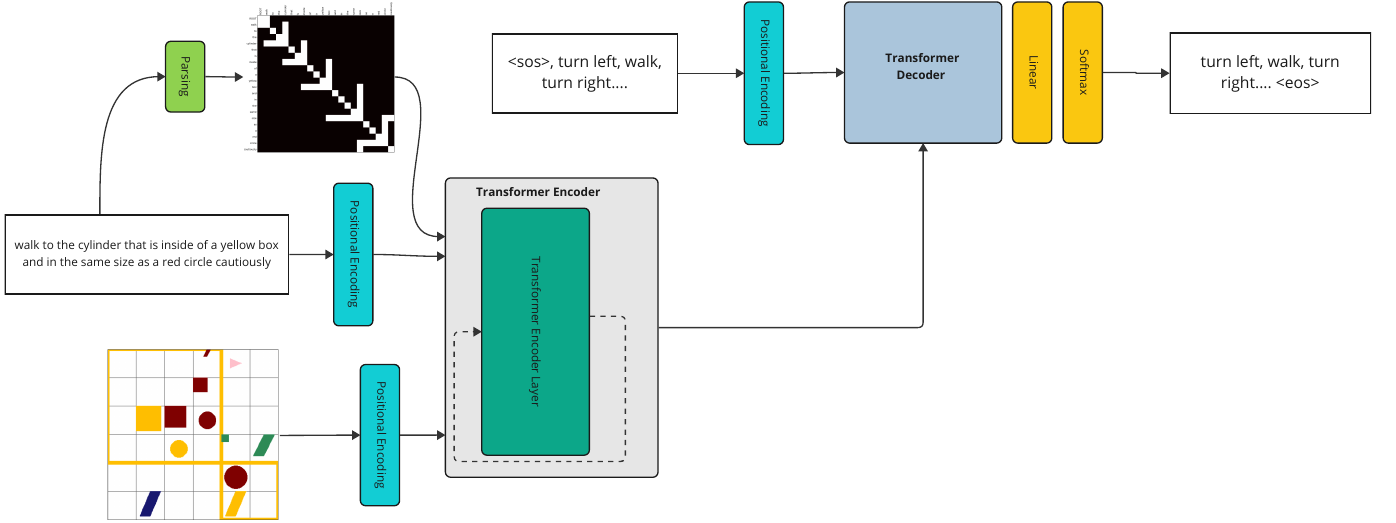}
    \caption{Overall architecture of the proposed model.}
    \label{fig:arch}
\end{figure*}

To address the challenge at hand, we implemented a multimodal transformer, as illustrated in Figure~\ref{fig:arch}. In this model, input commands are tokenized and then supplemented with positional encoding before passing to the transformer. Concurrently, the visual environment is segmented into 36 distinct cells, each serving as a visual token. After passing the visual token to a linear layer, these tokens receive positional encoding and are passed to the transformer.

We've employed a generic parser to seamlessly embed the structure of the textual modality into our model, thereby shaping attention masks for the encoder's textual self-attention. Prioritizing efficiency, parsing each input command is conducted during a preprocessing phase.

Our transformer is based on the GroCoT model~\cite{Sikarwar2022}. Each encoder layer employs a cross-attention mechanism between modalities, followed by modality-specific self-attention. Our computed input command masks are utilized in the self-attention modules of the textual modality. Remarkably, encoder layer weights are consistently shared across all layers. 

In the end, we concatenate the encoded result of each modality and pass it to the transformer's auto-regressive decoder to generate the action sequence corresponding to the input command given the environment.

\begin{figure}[t]
         \centering
         \includegraphics[width=0.47\textwidth]{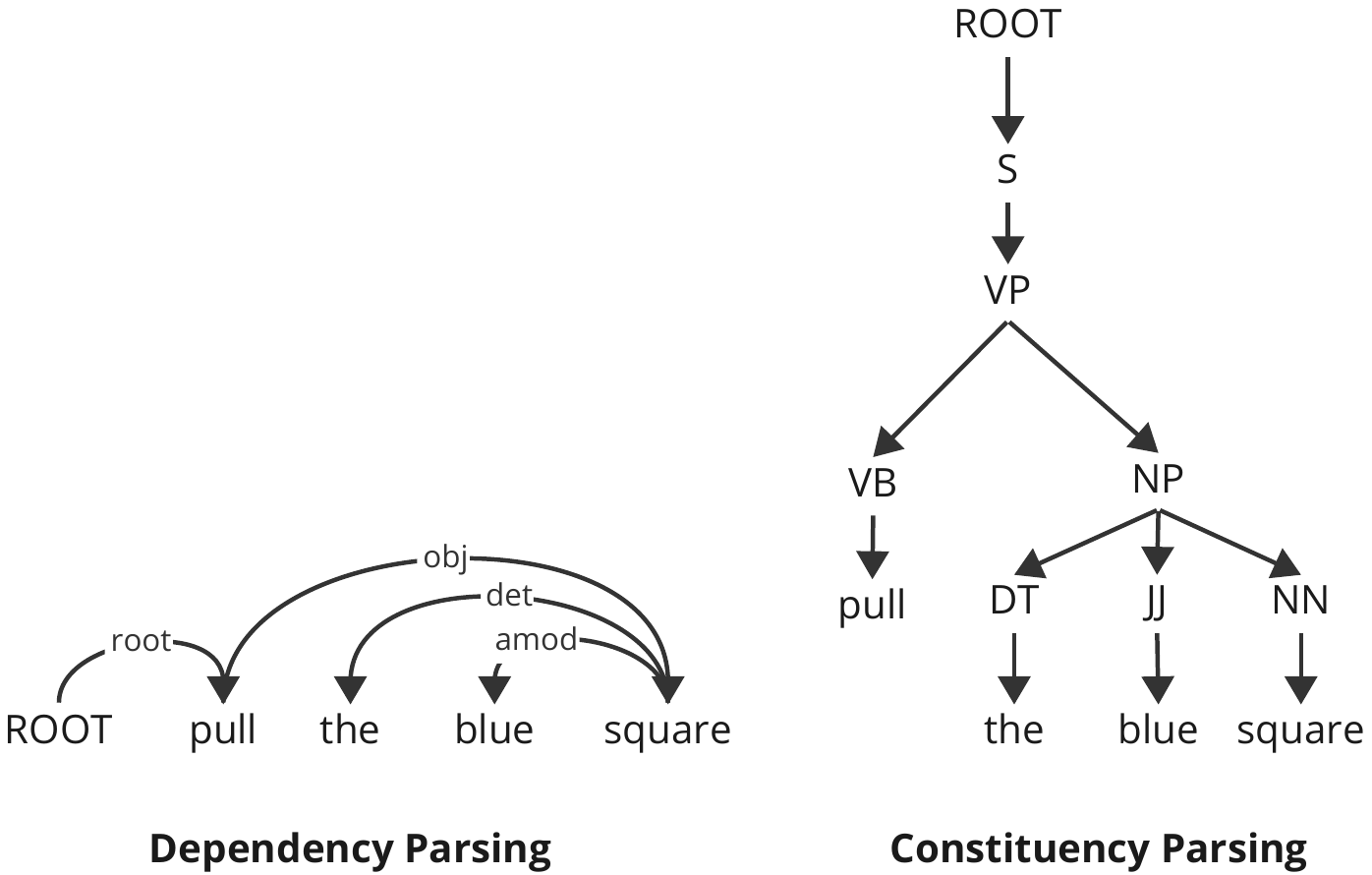}
     \caption{Examples of parse trees.}
     \label{fig:parse}
\end{figure}

\subsection{Syntax-guided attention}
One main component of our proposed model is exploiting the syntactic structure of the command. For this aim, we investigate using both dependency and constituency parsing.  Dependency and constituency trees can be used to analyze the grammatical structure of sentences. Dependency trees focus on the grammatical relationships between individual words, where each word except the root depends on another, and the edges of the tree signify these dependencies. However, constituency trees emphasize the hierarchical organization of words into larger syntactic units or constituents, with internal nodes representing these groupings and leaves representing individual words. While dependency trees are more concerned with identifying grammatical roles and relationships between words, constituency trees aim to show how words group together into larger syntactic units, often carrying syntactic labels like NP (noun phrase) or VP (verb phrase)  \cite{foscarin2023predicting, DBLP:conf/taln/HearneOT08}. Examples of these parse trees are shown in Figure \ref{fig:parse}.

\begin{figure}[h!]
     \centering
     \begin{subfigure}[b]{0.47\textwidth}
         \centering
         \includegraphics[width=\textwidth]{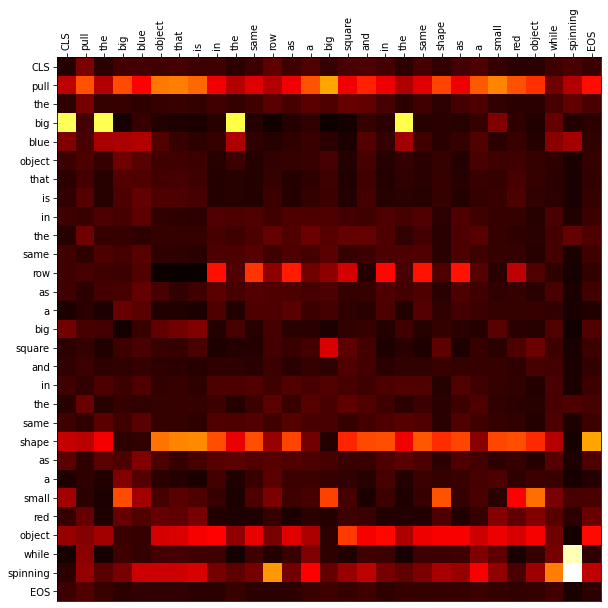}
         \caption{Self-attention w/o masking}
         \label{fig:self-without-mask}
     \end{subfigure}
     \hfill
     \begin{subfigure}[b]{0.47\textwidth}
         \centering
         \includegraphics[width=\textwidth]{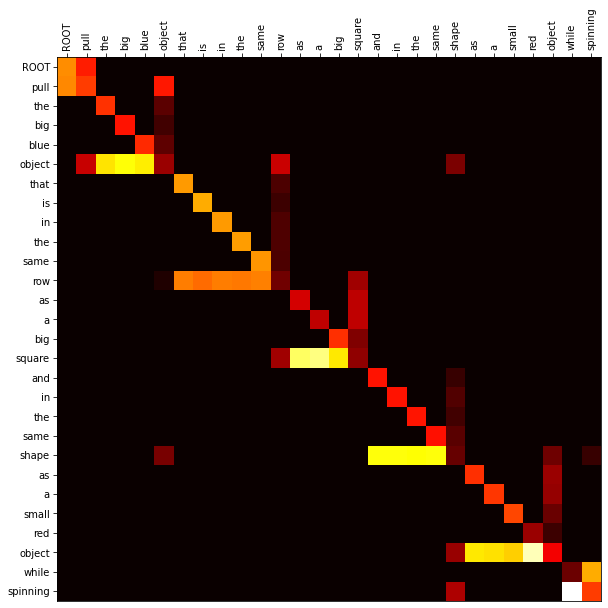}
         \caption{Self-Attention w/ masking}
         \label{fig:self-with-mask}
     \end{subfigure}
        \caption{Self-Attention example from the A2 test set of ReaSCAN dataset. Figures (a) and (b) depict the averaged self-attention map from our models' over all encoder layers and heads. Rows and columns correspond to text tokens. Brighter attention cells indicate higher attention weights}
        \label{fig:self-att}
\end{figure}

\textbf{Syntax-guided attention masking.} We use the syntactic information to guide the self-attention module of transformer encoder layers as depicted in Figures~\ref{fig:arch} and~\ref{fig:self-with-mask}. We force each token to only attend to the tokens connected in the syntax tree. In this way, we avoid faulty attention patterns and overfitting irrelevant parts of the sentence. In addition, by imposing the structure with a parse tree, our model can capture the nesting structure of the command's meaning and the relationships between its components. By making the structural information explicit, our model can potentially extrapolate the meaning of novel combinations and nesting linguistic structures encompassing higher reasoning depth.

\subsection{Weight Sharing}

Parameter sharing is a strategic approach where identical learned parameters are applied across various positions or layers within a model. This technique enables the reuse of the same encoder unit at each phase of the transformer encoder \cite{dehghani2019universal}. Such an approach not only streamlines the model but also nurtures the acquisition of more robust and adaptable representations of the input \cite{ontanon-etal-2022-making}.
 The findings of \citet{kim2021improving} demonstrate that a transformer employing attention masking requires extended training epochs for convergence, potentially due to masking-induced backpropagation constraints. In light of this, we hypothesize that introducing weight sharing might counterbalance this challenge. Weight sharing reduces the model's complexity by decreasing the number of parameters, which could lead to faster convergence. This method acts as a form of regularization, stabilizing training and facilitating smoother optimization landscapes.
In addition, \citet{ontanon-etal-2022-making} show that a transformer with shared weights across its encoder layers is arguably endowed with a more suitable inductive bias that allows the model to learn the primitive concepts. We hypothesize this will positively affect learning spatial relations or object-property relations, which are frequently used in our model's input. 
Motivated by these advantages, we incorporated this weight sharing technique into our transformer model to evaluate its efficacy in a multimodal setting. Beyond the enhanced generalizability, weight sharing serves as a computational benefit by reducing the number of learnable parameters during the training phase.

\section{Experiments}

\textbf{Implementation Details.} Our model architecture is founded on the GroCoT framework as detailed by \citet{Sikarwar2022} and is implemented using the PyTorch machine learning library \cite{pytorch}. Also, we employed the pre-trained stanza toolkit \cite{qi2020stanza} for constituency and dependency parsing. We used 48 GB A6000 GPUs accompanied by 756GB RAM. On average, each experiment took about 52 hours to train the models from scratch, with the Adam optimizer \cite{kingma2017adam} parameter updates throughout the training regimen. To ensure a rigorous evaluation, we used the same specialized compositional validation set as \citet{Sikarwar2022}, drawing 500 samples from each compositional division of the primary dataset. Model proficiency was assessed against this validation set, with the highest-performing model designated as our optimal choice. Our results are presented as an average derived from three independent runs, each initialized with a random seed. We ran the models for the ReaSCAN benchmark for 120 epochs, and the models for the gSCAN and GSRR benchmarks for 100 epochs. Hyperparameters used for the experiments of each dataset are shown in Appendix \ref{app:parmas}. The code and models proposed in this work are all available in GitHub\footnote{\url{https://github.com/HLR/Syntax-Guided-Transformers}}.

\textbf{Datasets. }
We used gSCAN \cite{ruis2020benchmark}, GSRR \cite{qiu-etal-2021-improving}, ReaSCAN \cite{Wu-ReaSCAN} benchmarks for evaluation.  
The Grounded SCAN (gSCAN) dataset is a benchmark tailored for examining compositional generalization in machine learning models by translating natural language commands into actions in a grid-world scenario. Its unique splits ensure models move beyond rote memorization to deep compositional understanding of concepts. The Grounded Systematic Relation Reasoning (GSRR) dataset extends gSCAN by aligning natural language instructions intricately with visual elements, emphasizing spatial relationships and object references. ReaSCAN, a further development, brings the challenges of real-world reasoning into this environment by introducing more challenging tasks and concept relations. Together, these datasets offer a high-complexity framework for assessing the compositional and relational understanding of machine learning models in visual environments. A detailed explanation of both the gSCAN and Grounded Systematic Relation Reasoning datasets can be found in Appendix \ref{app:dataset}.

\textbf{Baselines.} We embarked on a series of experiments designed to evaluate our model's effectiveness compared to the most recent state-of-the-art models on the mentioned multimodal compositional generalization datasets. We include the following baselines. 
\textbf{(a)}~\citet{ruis2020benchmark} (Multimodal LSTM)  is a fusion of sequence-to-sequence (seq2seq) architecture with a visual encoder, employing a recurrent 'command encoder' to process the instructions. \textbf{(b)}~\citet{gao-etal-2020-systematic} (GCN-LSTM) integrates a Graph Convolutional Neural (GCN) network with a multimodal LSTM. The command encoding is achieved via a BiLSTM equipped with multi-step textual attention, while the world is encoded through a GCN layer. 
\textbf{(c)}~\citet{qiu-etal-2021-improving} (Multimodal Transformer) is a multimodal transformer equipped with cross-attention for multimodal compositional generalization. 
\textbf{(d)}~\citet{Sikarwar2022} (GroCoT) is another transformer-based model that incorporates interleaved self-attention into the multimodal transformer with cross-attention.

\begin{table*}[h]
    \centering

    \begin{tabular}{l|c|c|c|c|c|c|c|c}
\hline
Model & A1 & A2 & A3& B1 & B2 & C1 & C2 & Avg \\
\hline

\footnotesize LSTM*  & \footnotesize  50.4 &\footnotesize 14.7 &\footnotesize 50.9 &\footnotesize 52.2 &\footnotesize 39.4 &\footnotesize 49.7 &\footnotesize 25.7 &\footnotesize 40.40 \\
\footnotesize  GCN-LSTM &\footnotesize 92.3 &\footnotesize 42.1 &\footnotesize 87.5 &\footnotesize 69.7 &\footnotesize 52.8 &\footnotesize 57.0 &\footnotesize 22.1 &\footnotesize 60.50 \\
\footnotesize  Transformer*   &\footnotesize 96.7 &\footnotesize 58.9 &\footnotesize 93.3 &\footnotesize 79.8 &\footnotesize 59.3 &\footnotesize 75.9 &\footnotesize 25.5 &\footnotesize 69.90\\
\footnotesize  GroCoT  &\footnotesize  99.6 &\footnotesize 93.1 &\footnotesize 98.9 &\footnotesize 93.9 &\footnotesize 86.0 &\footnotesize 76.3 &\footnotesize \textbf{27.3} &\footnotesize 82.2 \\
\footnotesize Constituency$^\dagger$  & \footnotesize \textbf{99.75}\tiny $\pm$0.11    & \footnotesize 96.70\tiny $\pm$1.40    & \footnotesize \textbf{99.68}\tiny $\pm$0.10    & \footnotesize 95.19\tiny $\pm$1.17    & \footnotesize 88.37\tiny $\pm$1.50    & \footnotesize 69.07\tiny $\pm$0.60    & \footnotesize 27.00\tiny $\pm$0.54  & \footnotesize  82.25\tiny $\pm$0.63\\

\footnotesize Dependency$^\dagger$  & \footnotesize \footnotesize 99.65\tiny $\pm$0.9    & \footnotesize \textbf{97.37}\tiny $\pm$0.48    & \footnotesize 99.62\tiny $\pm$0.07    & \footnotesize \textbf{95.46}\tiny $\pm$2.01    & \footnotesize \textbf{90.15}\tiny $\pm$3.88    & \footnotesize \textbf{92.55}\tiny $\pm$1.51    & \footnotesize 21.77\tiny $\pm$5.25    & \footnotesize \textbf{85.22}\tiny $\pm$0.87\\
\hline

    \end{tabular}
    \caption{The result of our proposed model on the ReaSCAN dataset test splits. The results are an average of three runs. $\dagger$ denotes the models with masking. Models marked with * refer to the multimodal version of their implementation.}
    \label{tab:reascan-result}
\end{table*}

\begin{table*}[h]
    \centering
    \begin{tabular}{l|c|c|c|c|c|c|c}
\hline
Model & A & B & C&  E & F  & H & Comp. Avg \\
\hline
\footnotesize LSTM* & \footnotesize 97.7 & \footnotesize 54.9& \footnotesize 23.5 & \footnotesize 35.0 & \footnotesize 92.5 &  \footnotesize 22.7 & \footnotesize 32.7 \\
\footnotesize GCN-LSTM  & \footnotesize 98.6 & \footnotesize 99.1 & \footnotesize 80.3 & \footnotesize 87.3 & \footnotesize 99.3 & \footnotesize \textbf{33.6} & \footnotesize - \\
\footnotesize Transformer*  & \footnotesize 99.9 & \footnotesize 99.9 & \footnotesize 99.3  & \footnotesize 99.0 & \footnotesize 99.9  & \footnotesize 22.2 & \footnotesize 60.0\\
\footnotesize GroCoT  & \footnotesize  99.9 & \footnotesize 99.9 & \footnotesize 99.9 & \footnotesize 99.8 & \footnotesize 99.9 &  \footnotesize 22.9 & \footnotesize 60.4 \\
\footnotesize \footnotesize Constituency$^\dagger$  & \footnotesize \textbf{99.95}\tiny $\pm$0.07    & \footnotesize \textbf{99.92}\tiny $\pm$0.06    & \footnotesize \textbf{99.88}\tiny $\pm$0.11      & \footnotesize 99.88\tiny $\pm$0.09  &  \footnotesize \textbf{100.00}\tiny $\pm$0.00 & \footnotesize 22.84\tiny $\pm$0.93    & \footnotesize 60.36\tiny $\pm$0.11 \\
\footnotesize Dependency$^\dagger$ & \footnotesize  99.92\tiny $\pm$0.09    & \footnotesize 99.85\tiny $\pm$0.18    & \footnotesize 99.86\tiny $\pm$0.11    &  \footnotesize \textbf{99.96}\tiny $\pm$0.06    & \footnotesize 
99.89\tiny $\pm$0.16 & \footnotesize 23.89\tiny $\pm$1.54    & \footnotesize \textbf{60.49}\tiny $\pm$0.20 \\

\hline

    \end{tabular}
    \caption{The result of our proposed model on the gSCAN dataset test splits. The results are an average of three runs. We did not report the results on D and G splits since we achieved 0.00\tiny$\pm$0.00 \normalsize \% performance, But take them into account in the averaged result. $\dagger$ denotes the models with masking. Models marked with * refer to the multimodal version of their implementation.}
    \label{tab:gscan-results}
\end{table*}

\begin{table*}[h]
    \centering
    
    \begin{tabular}{l|c|c|c|c|c|c|c}
\hline

 Model & I & II & III& IV & V & VI & Comp. Avg \\
\hline

\footnotesize LSTM*  & \footnotesize 86.5 & \footnotesize 40.1 & \footnotesize 86.1 & \footnotesize 5.5 & \footnotesize  81.4 & \footnotesize 81.8 & \footnotesize 58.9 \\
\footnotesize Transformer*   & \footnotesize 94.7 & \footnotesize  64.4 & \footnotesize 94.9 &  \footnotesize 49.6 & \footnotesize 59.3 & \footnotesize 49.5 & \footnotesize 63.5 \\
\footnotesize GroCoT  &  \footnotesize 99.9 & \footnotesize 98.6 & \footnotesize \textbf{99.9} & \footnotesize 99.7 & \footnotesize \textbf{99.5} & \footnotesize 96.5 & \footnotesize 98.8 \\
\footnotesize Constituency$^\dagger$ & \footnotesize  99.85\tiny $\pm$0.00   & \footnotesize 99.90\tiny $\pm$0.03    & \footnotesize 99.16\tiny $\pm$0.26    & \footnotesize 99.88\tiny $\pm$0.03    & \footnotesize 96.73\tiny $\pm$2.16    & \footnotesize \textbf{97.85}\tiny $\pm$0.46    & \footnotesize  98.58\tiny $\pm$0.39 \\
\footnotesize Dependency$^\dagger$ & \footnotesize \textbf{99.91}\tiny $\pm$0.02    & \footnotesize \textbf{99.93}\tiny $\pm$0.01    & \footnotesize 99.41\tiny $\pm$0.28    & \footnotesize \textbf{99.96}\tiny $\pm$0.01    & \footnotesize 99.03\tiny $\pm$0.23    & \footnotesize 97.38\tiny $\pm$0.63  & \footnotesize \textbf{99.07}\tiny $\pm$0.16 \\
\hline
    \end{tabular}
    \caption{The result of our proposed model on the GSRR dataset test splits. The results are an average of three runs. $\dagger$ denotes the models with masking. Models marked with * refer to the multimodal version of their implementation.}
    \label{tab:google-results-full}
\end{table*}

\textbf{Results. }We comprehensively evaluated our approach across all the previously mentioned benchmarks compared to the baselines. 
Alongside the accuracy and efficacy metrics, we also provide insights into the computational overhead associated with our method. Furthermore, a qualitative analysis is presented, delving deeper into our approach's performance nuances and strengths.

The benchmark results, presented in Tables \ref{tab:reascan-result}, \ref{tab:gscan-results}, and \ref{tab:google-results-full}, demonstrate our model's superior performance over all reported models, with a notable 3\% improvement on the average of ReaSCAN benchmark splits. This substantiates our hypothesis that incorporating syntactic parsing significantly boosts the model's generalization derived from grounded compositional training data. Moreover, dependency parsing consistently outperformed constituency or marked a very similar performance across multiple benchmarks, including GSRR and gSCAN. Our model displayed improvements across nearly all ReaSCAN splits except for C2. As per \citet{Sikarwar2022}, the C2 split is "unfair," lacking the required information in training data for comprehensive model training. Even including syntactic information could not improve the model's performance on this split and even caused a decrease in the performance. Our methodology also showcased its merit in the object property test cases (A1-3), effectively constraining attention to words pertinent to target object descriptors. For instance, as shown in Figure \ref{fig:self-att}, the attention weights from the properties to the corresponding objects are high.

Notably, our model exhibited considerable strides in the C1 split, indicative of the value added by syntactic information. For a more reliable comparison, we applied a t-test to our C1 test split results. Using a significance level ($\alpha$) of 0.05, this statistical analysis provided further validation for the observed enhancements in our model's performance, particularly within the context of the C1 test split. Furthermore, our model exhibits enhanced performance on the GSRR dataset. As illustrated in Table \ref{tab:google-results-full}, both variants of our model demonstrate improvements in the II split. It is worth noting that the II split shares the same challenge as the A2 split from the ReaSCAN dataset but in a less complex environment. 

While our proposed techniques effectively address splits A, B, C, E, and F, mirroring the successes of previous works such as \cite{Sikarwar2022} and \cite{qiu-etal-2021-improving}, they struggle with challenges presented by specific gSCAN compositionality splits, notably D, G, and H. These particular splits are designed to assess the model's capacity for systematic generalization when novel patterns should occur on the output sequence rather than in grounding the input instruction~\cite{Sikarwar2022}, a facet that is not expected to be captured by our proposed model.

\begin{table*}[h]
    \centering
    \begin{tabular}{c|c|c|c|c|c|c|c|c|c}
\hline
    
\footnotesize W/S &  \footnotesize Mask &  A1 & A2 & A3& B1 & B2 & C1 & C2 & Avg \\

\hline
 - & \footnotesize - & \footnotesize  99.29\tiny $\pm$0.27    & \footnotesize 91.82\tiny $\pm$6.50    & \footnotesize 98.49\tiny $\pm$1.17    & \footnotesize 93.50\tiny $\pm$0.85    & \footnotesize 83.15\tiny $\pm$1.41    & \footnotesize 75.85\tiny $\pm$1.35    & \footnotesize \textbf{25.03}\tiny $\pm$6.82  & \footnotesize  81.02\tiny $\pm$0.22 \\

 \checkmark & \footnotesize - & \footnotesize 99.68\tiny $\pm$0.22    & \footnotesize 97.09\tiny $\pm$1.72    & \footnotesize 99.64\tiny $\pm$0.20    & \footnotesize 94.86\tiny $\pm$0.77    & \footnotesize 81.49\tiny $\pm$4.27    & \footnotesize 66.30\tiny $\pm$6.65    & \footnotesize 21.66\tiny $\pm$1.83 & \footnotesize  80.10\tiny $\pm$1.08 \\

-& \footnotesize \footnotesize Dep. & \footnotesize 98.09\tiny $\pm$0.27    & \footnotesize 85.21\tiny $\pm$6.85    & \footnotesize 97.35\tiny $\pm$0.75    & \footnotesize 93.61\tiny $\pm$2.75    & \footnotesize 90.62\tiny $\pm$1.59    & \footnotesize 75.27\tiny $\pm$1.77    & \footnotesize 21.91\tiny $\pm$1.63  & \footnotesize  80.29\tiny $\pm$1.43\\
\checkmark& \footnotesize Dep. & \footnotesize \textbf{99.65}\tiny $\pm$0.9    & \footnotesize \textbf{97.37}\tiny $\pm$0.48    & \footnotesize \textbf{99.62}\tiny $\pm$0.07    & \footnotesize \textbf{95.46}\tiny $\pm$2.01    & \footnotesize \textbf{90.15}\tiny $\pm$3.88    & \footnotesize \textbf{92.55}\tiny $\pm$1.51    & \footnotesize 21.77\tiny $\pm$5.25    & \footnotesize \textbf{85.22}\tiny $\pm$0.87\\

\hline

    \end{tabular}
    \caption{The ablation study result of our modifications on ReaSCAN dataset test splits. Results are reported on an average of three runs. We evaluate every combination of components from our best model. W/S stands for weight sharing, and the \checkmark shows the presence of the module. \textit{Dep} in this table refers to the Dependency masking. We evaluate the model with or without dependency masking in the masking part.}
    \label{tab:reascan-ablation}
\end{table*}

\begin{figure*}[h!]
     \centering
     \begin{subfigure}[b]{0.3\textwidth}
         \centering
        \includegraphics[width=1\textwidth]{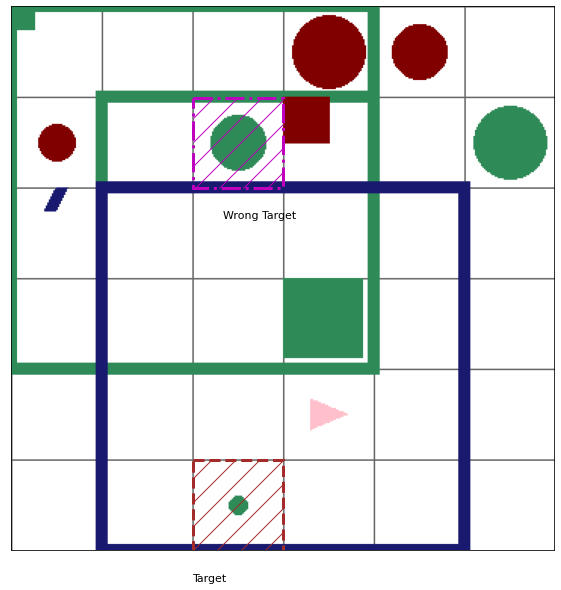}
         \caption{Environment}
         \label{fig:env}
     \end{subfigure}
     \hfill
     \begin{subfigure}[b]{0.3\textwidth}
         \centering
         \includegraphics[width=1.035\textwidth]{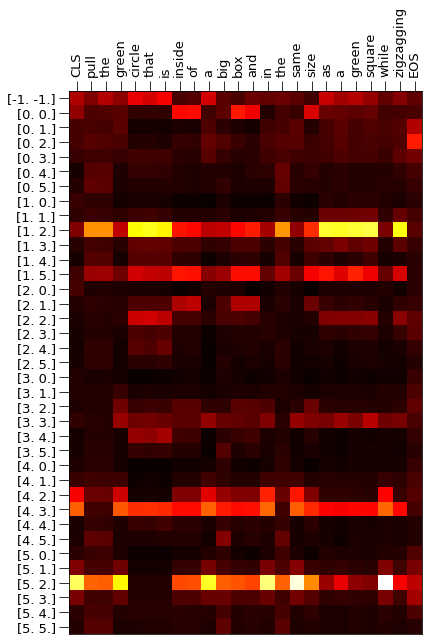}
         \caption{Cross-attention w/o masking}
         \label{fig:without-mask}
     \end{subfigure}
     \hfill
     \begin{subfigure}[b]{0.3\textwidth}
         \centering
         \includegraphics[width=\textwidth]{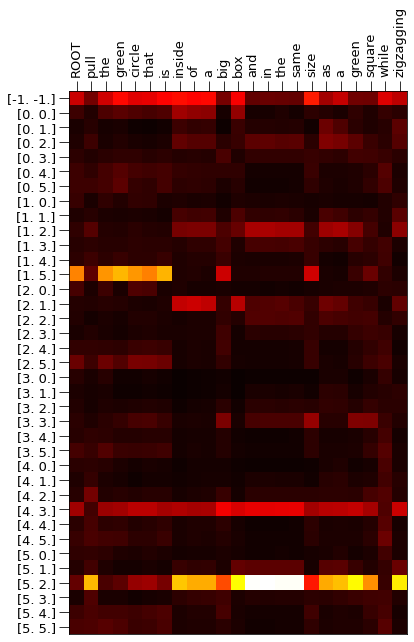}
         \caption{Cross-Attention w/ masking}
         \label{fig:with-mask}
     \end{subfigure}
        \caption{Cross-Attention from Text-to-Image. In Figure (a), the purple zone indicates the model's incorrect object selection, while the red zone highlights the accurate choice. Figures (b) and (c) depict the averaged cross-attention map from our models over encoder layers and attention heads. The rows represent environment cells (the first element shows the row, and the second shows the column index, both starting from 0), and the columns correspond to text tokens. Brighter attention cells signify elevated attention weights.}
        \label{fig:cross-att}
\end{figure*}
\subsection{Ablation }
For a granular understanding of the contributions from each alteration to the baseline model, we undertook an ablation study. This involved the sequential removal of each modification to measure its individual impact. 
As depicted in Table \ref{tab:reascan-ablation}, while individual modifications did not significantly change the baseline, their collective integration enhanced the model's generalization. Remarkably, eliminating dependency parsing or weight sharing resulted in a noticeable performance dip. The improvement upon integration posits that weight sharing can potentially offset the masking prolonged convergence challenge by reducing parameter count, thereby mitigating the convergence issues.

\subsection{Qualitative Analysis}
In our previous discussions, we highlighted the significance of integrating dependency parsing as a fundamental approach to understanding the complex structures inherent in sentences. This integration is not a mere enhancement; it critically enriches the model's grounding capabilities, offering a more robust bridge between raw textual sequences and their semantic structure.

To provide empirical evidence of our technique for guiding attention, we conducted an analysis of the cross-attention module. We aimed to compare its behavior before and after applying attention masking. The results, presented in Figure \ref{fig:cross-att}, indicate a clear trend: in 86\% of validation samples, the cross-attention module exhibits a pronounced focus on the target object.


Figures \ref{fig:without-mask} and \ref{fig:with-mask} elucidate the impact of self-attention masking on these weights. After using attention masking (see Figure \ref{fig:without-mask}), the attention distribution becomes notably sparser; instead of individual words attending in isolation to every potentially relevant cell, they now form cohesive compositional expressions, each attending to the corresponding cells as a whole. For instance, in Figure \ref{fig:with-mask}, "and in the same," phrase's tokens attend to cells (1,2), (4,3), and (5,2) together with greater attention on the target object in contrast to their attention pattern without masking. 


\begin{table}[t]
\centering
\begin{tabular}{l|c}
\hline
\textbf{Model} & \textbf{\#Parameters} \\
\hline
Multimodal LSTM & 74K \\
Multimodal Transformer & 3M \\
GroCoT & 4.6M \\
Dependency$^\dagger$ (ours) & 1.9M \\
\hline
\end{tabular}
\caption{Comparing model parameters: our model vs. current state-of-the-art models. Dependency$^\dagger$ refers to the model with dependency parsing for attention masking.}
\label{tab:parameter_comparison}
\end{table}
\subsection{Efficiency Analysis}

In the realm of modern model design, the challenge lies in amplifying capabilities while managing computational overhead. Our methodology adeptly navigates this balance. A cornerstone of our model's efficiency is the strategic adoption of weight sharing within the transformer encoder. By reusing weights across different components, we significantly reduce the parameter space. This not only streamlines memory utilization and accelerates training but also acts as an implicit regularizer, bolstering the model's generalization capabilities and reducing overfitting. Further enhancing this is our implementation of attention masking, which refines computational efficiency. By enabling the model to selectively bypass attention to certain tokens, we can optimize the model to avoid redundant computational processes, ensuring optimal resource allocation and superior performance.

As illustrated in Table \ref{tab:parameter_comparison}, our model stands out in terms of efficiency. Despite having fewer parameters (1.9M) than the models by \citet{qiu-etal-2021-systematic} and \citet{Sikarwar2022}, which have 3M and 4.6M parameters respectively, our model consistently outperforms them across all benchmarks.

\section{Conclusion}
Our research demonstrated that exploiting the syntactic structure of compositional and complex linguistic and spatial expressions improved the grounding ability of the instruction-follower agent in multimodal environments.
Our results indicated improvements compared to the previous state-of-the-art models. In particular, we show that our proposed model is effective for generalization on tasks and test splits that require generalization over unobserved reasoning depths, such as the C1 split in the ReaSCAN dataset. By utilizing the syntactic-guided attention masking along with the weight sharing, we achieved not only more accurate but also more parameter-efficient models for grounded compositional generalization.

\section*{Limitations}

Despite the promising results achieved in our study, several limitations warrant consideration:

    \textbf{Synthetic Data:} Our experiments predominantly rely on synthetic datasets. While these datasets provide a controlled environment for assessing model performance, they might not capture the complexities and nuances of real-world data. Evaluating the models on real-world datasets is crucial to ensure their practical applicability.

    \textbf{Error Propagation from the Parser:} The model's performance is intrinsically tied to the accuracy of the pre-trained parsers we utilized. Errors or inaccuracies in parsing can lead to suboptimal model outputs. Additionally, our synthetic data, being unambiguous, might not reveal the full extent of potential parser-related issues.

    \textbf{Computational Constraints:} Due to computational limitations, the hyperparameter search might not have been exhaustive. A more comprehensive exploration might yield better model configurations.

\section*{Acknowledgement}

This project is supported by the National Science Foundation (NSF) CAREER award 2028626 and partially supported by the Office of Naval Research
(ONR) grant N00014-20-1-2005. Any opinions,
findings, and conclusions or recommendations expressed in this material are those of the authors and
do not necessarily reflect the views of the National
Science Foundation nor the Office of Naval Research. We thank all reviewers for their thoughtful
comments and suggestions.

\bigskip

\bibliography{anthology,custom}
\bibliographystyle{acl_natbib}

\appendix

\section{Hyperparameters}
\label{app:parmas}
Here, we present the hyperparameters used in the models for every benchmark in Table \ref{tab:hyperparams}.
\begin{table*}[t]
    \centering
    \begin{tabular}{l|c|c|c}
    \hline
    Hyperparameter & gSCAN & GSRR & ReaSCAN \\
    \hline
         Number of Vision Self-Attention Layers& $3$ & $3$ & $6$   \\
         Number of Text Self-Attention Layers& $3$ & $3$ & $6$  \\
         Number of Cross-Attention Layers & $3$ & $3$ & $6$  \\
         Number of Decoder Layers&  $3$ & $3$ & $6$ \\
         Embedding Size & $128$ &$128$ & $128$ \\
         Hidden Layer Size & $256$ & $256$ & $256$ \\
         Number of Attention Heads & $8$ &$8$ & $8$ \\
         Learning Rate & $5\times 10^{-5}$ & $1\times 10^{-5}$ & $1\times 10^{-5}$ \\
         Batch Size & $64$& $64$& $32$ \\
         Dropout & $0.1$ & $0.1$ & $0.1$ \\
         Number of Epochs & $100$ & $100$ & $120$ \\
         \hline
    \end{tabular}
    \caption{Hypterparameters used in the experiments.}
    \label{tab:hyperparams}
\end{table*}

\section{Datasets Description}
\label{app:dataset}

\subsection{Grounded SCAN dataset}
\label{app:gscan}
The Grounded SCAN (gSCAN) dataset is a pivotal benchmark for assessing compositional generalization in machine learning models. Evolving from the foundational SCAN \cite{lake2018generalization} dataset, gSCAN is designed to evaluate a model's proficiency in translating command sequences into actions within a grid world environment, emphasizing on compositional challenges.

This benchmark offers systematic test splits that rigorously examine a model's capability to generalize beyond its training data. These compositional splits include:

\begin{itemize}
    \item \textbf{A (Random)}: Random data with a similar distribution to the training data.
    \item \textbf{B (Color-Shape)}: Novel composition of object properties in the testing. Yellow squares are referred to by color and shape.
    \item \textbf{C (Color Only)}: Red squares as target.
    \item \textbf{D (Novel Direction)}: Challenges a model's spatial comprehension, with targets set in unfamiliar directions, the southwest.
    \item \textbf{E (Novel Contextual References)}: Evaluates a model's understanding of relative sizes, with commands pointing to circles of size 2 described as "small."
    \item \textbf{F (Novel Composition of Actions and Arguments)}: Probes a model's grasp of object classes and their nuances, exemplified by squares of size 3 necessitating two pushes.
    \item \textbf{G (Adverb)}: Commands carrying the adverb "cautiously" test how well the model interprets action modifiers after seeing limited training samples (k=1).
    \item \textbf{H (Adverb-Verb Combination)}: Generalizes to commands pairing actions and their modifiers, like "while spinning" combined with "pull."
\end{itemize}

The compositional test splits of the gSCAN dataset ensure that models are not indulging in learning statistical shortcuts but are genuinely mastering compositional reasoning. In gSCAN, every command is mapped to an action sequence for an agent in the grid world, whether moving to a particular spot or interacting with a distinct described object.

\subsection{Grounded Systematic Relation Reasoning dataset (GSRR)}
\label{app:GSRR}

The Grounded Systematic Relation Reasoning (GSRR) dataset, introduced by \cite{qiu-etal-2021-improving}, extends the gSCAN benchmark. Their initial analyses of the gSCAN dataset indicated its efficacy; the authors observed that several remaining challenges might not be primarily tied to visual grounding. In light of this, they proposed the GSRR task, characterized by an elevated complexity in aligning natural language instructions with the visual environment.

In this dataset, language expressions specifically delineate target objects and explicitly describe their relationships with a secondary referenced object. They incorporate two types of relations into our dataset: immediate adjacency ("next to") and cardinal directions such as "north" and "west." In addition, they put visual distractors objects within the environment to emphasize the critical role of spatial relations in identifying the target objects.

The dataset is systematically divided into various splits to ensure a comprehensive assessment:

\begin{itemize}
\item \textbf{ I (Random)}: Similar distribution as the training.
\item \textbf{II (Visual)}: Commands centering on "red squares" either as targets or references.
\item \textbf{III (Relation)}: Complex instructions involving combinations like "green squares" and "blue circles."
\item \textbf{IV (Referent)} Emphasizing "yellow squares" as primary targets.
\item \textbf{V (Relative Position 1)}: Commands where targets are situated to the "north" of their reference points.
\item \textbf{VI (Relative Position 2)}: Instructions where targets are located "southwest" relative to their references.
\end{itemize}

\section{Evaluation Card}
Here, we present the evaluation card of our compositional generalization experiments based on \cite{hupkes2023stateoftheart} taxonomy.
\newcommand{\tabularwidth}{\columnwidth}

\newcommand{\expone}{$\square$}
        
\renewcommand{\arraystretch}{1.1}         
\setlength{\tabcolsep}{0mm}         
\begin{tabular}{|p{\tabularwidth}<{\centering}|}         
\hline
               
\rowcolor{gray!60}               
\textbf{Motivation} \\               
\footnotesize
\begin{tabular}{p{0.25\tabularwidth}<{\centering} p{0.25\tabularwidth}<{\centering} p{0.25\tabularwidth}<{\centering} p{0.25\tabularwidth}<{\centering}}                        
\textit{Practical} & \textit{Cognitive} & \textit{Intrinsic} & \textit{Fairness}\\
\expone\hspace{0.8mm}		
& 		
& \expone\hspace{0.8mm}		
& 		

\vspace{2mm} \\
\end{tabular}\\
               
\rowcolor{gray!60}               
\textbf{Generalisation type} \\               
\footnotesize
\begin{tabular}{m{0.17\tabularwidth}<{\centering} m{0.20\tabularwidth}<{\centering} m{0.14\tabularwidth}<{\centering} m{0.17\tabularwidth}<{\centering} m{0.18\tabularwidth}<{\centering} m{0.14\tabularwidth}<{\centering}}                   
\textit{Compo- sitional} & \textit{Structural} & \textit{Cross Task} & \textit{Cross Language} & \textit{Cross Domain} & \textit{Robust- ness}\\
\expone\hspace{0.8mm}		
& \expone\hspace{0.8mm}		
& 		
& 		
& 		
& 		

\vspace{2mm} \\
\end{tabular}\\
             
\rowcolor{gray!60}             
\textbf{Shift type} \\             
\footnotesize
\begin{tabular}{p{0.25\tabularwidth}<{\centering} p{0.25\tabularwidth}<{\centering} p{0.25\tabularwidth}<{\centering} p{0.25\tabularwidth}<{\centering}}                        
\textit{Covariate} & \textit{Label} & \textit{Full} & \textit{Assumed}\\  
& 		
& \expone\hspace{0.8mm}		
& 		

\vspace{2mm} \\
\end{tabular}\\
             
\rowcolor{gray!60}             
\textbf{Shift source} \\             
\footnotesize
\begin{tabular}{p{0.25\tabularwidth}<{\centering} p{0.25\tabularwidth}<{\centering} p{0.25\tabularwidth}<{\centering} p{0.25\tabularwidth}<{\centering}}                          
\textit{Naturally occuring} & \textit{Partitioned natural} & \textit{Generated shift} & \textit{Fully generated}\\
& 		
& 		
& \expone\hspace{0.8mm}		

\vspace{2mm} \\
\end{tabular}\\
             
\rowcolor{gray!60}             
\textbf{Shift locus}\\             
\footnotesize
\begin{tabular}{p{0.25\tabularwidth}<{\centering} p{0.25\tabularwidth}<{\centering} p{0.25\tabularwidth}<{\centering} p{0.25\tabularwidth}<{\centering}}                         
\textit{Train--test} & \textit{Finetune train--test} & \textit{Pretrain--train} & \textit{Pretrain--test}\\
\expone\hspace{0.8mm}		
& 		
& 		
& 		

\vspace{2mm} \\
\end{tabular}\\

\hline
\end{tabular}

\end{document}